# Flexible Modeling of Latent Task Structures in Multitask Learning

**Alexandre Passos**[†] APASSOS@CS.UMASS.EDU
Computer Science Department, University of Massachusetts, Amherst, MA USA

**Piyush Rai**[†] PIYUSH@CS.UTAH.EDU
School of Computing, University of Utah, Salt Lake City, UT USA

**Jacques Wainer** WAINER@IC.UNICAMP.BR
Information Sciences Department, University of Campinas, Brazil

**Hal Daumé III** HAL@UMIACS.UMD.EDU
Computer Science Department, University of Maryland, College Park, MD USA

## Abstract

Multitask learning algorithms are typically designed assuming some fixed, *a priori* known latent structure shared by all the tasks. However, it is usually unclear what type of latent task structure is the most appropriate for a given multitask learning problem. Ideally, the "right" latent task structure should be *learned* in a data-driven manner. We present a flexible, nonparametric Bayesian model that posits a mixture of factor analyzers structure on the tasks. The nonparametric aspect makes the model expressive enough to subsume many existing models of latent task structures (e.g, mean-regularized tasks, clustered tasks, low-rank or linear/non-linear subspace assumption on tasks, etc.). Moreover, it can also learn more general task structures, addressing the shortcomings of such models. We present a variational inference algorithm for our model. Experimental results on synthetic and real-world datasets, on both regression and classification problems, demonstrate the effectiveness of the proposed method.

## 1. Introduction

Learning problems do not exist in a vacuum. Often one is tasked with developing not one, but many classifiers for different tasks. In these cases, there is often not enough data to learn a good model for each task individually—real-world examples are prioritizing email messages across many users' inboxes (Aberdeen et al., 2011) and recommending items to users on web sites (Ning & Karypis, 2010). In these settings it is advantageous to transfer or share information across tasks. Multitask learning (MTL) (Caruana, 1997) encompasses a range of techniques to share statistical strength across models for various tasks and allows learning even when the amount of labeled data for each individual task is very small. Most MTL methods achieve this improved performance either by assuming some notion of similarity across tasks—for example, that all task parameters are drawn from a shared Gaussian prior (Chelba & Acero, 2006), have a cluster structure (Xue et al., 2007; Jacob & Bach, 2008), live on a low-dimensional subspace (Rai & Daumé III, 2010), share feature representations (Argyriou et al., 2007), or by modeling the task covariance matrix (Bonilla et al., 2007; Zhang & Yeung, 2010). Choosing the correct notion of task relatedness is crucial to the effectiveness of any MTL method. Incorrect assumptions can hurt performance and it is desirable to have a flexible model that can automatically *adapt* its assumptions for a given problem.

Motivated by this, we propose a nonparametric Bayesian MTL model by representing the task parameters (e.g., the weight vectors for logistic regression models) as being generated from a nonparametric mixture of nonparametric factor analyzers. Parameters are shared only between tasks in the same cluster and, within each cluster, across a linear subspace that regularizes what is shared. Moreover, by virtue of this being a nonparametric model, various existing MTL models result as special cases of our model; for example, the weight vectors are drawn from a single shared Gaussian prior, or form clusters (equivalently, gener-

[†]Contributed equally

Appearing in *Proceedings of the 29$^{th}$ International Conference on Machine Learning*, Edinburgh, Scotland, UK, 2012.

ated from a mixture of Gaussians), or live close to a subspace, etc. Our model can automatically interpolate between these assumptions as needed, providing the best fit to the given MTL problem.

In addition to offering a general framework for multitask learning, our proposed model also addresses several shortcomings of commonly used MTL models. For example, task clustering (Xue et al., 2007), which fits a full-covariance Gaussian mixture model over the weight vectors, is prone to overfitting on high dimensional problems as the number of learning tasks is usually much smaller than the dimensionality, making it difficult to estimate the covariance matrix. A model based on mixtures of factor analyzers, like ours, can deal with this issue by adaptively estimating the dimensionality of each component, using less parameters than in the full rank case. Likewise, models based on task subspaces (Zhang et al., 2006; Rai & Daumé III, 2010; Agarwal et al., 2010) assume that the weight vectors of all the tasks live on or close to a *single* shared subspace, which is known to lead to negative transfer in the presence of outlier tasks. Our model, based on a mixture of subspaces, circumvents these issues by allowing different groups of weight vectors to live in different subspaces when grouping all together them would not fit the data well. One can also view our model as allowing the sharing of statistical strengths at two levels: (1) by exploiting the cluster structure, and (2) by additionally exploiting the subspace structure *within* each cluster.

## 2. Background

In the context of MTL, since the task relatedness structure is usually unknown, the standard solution is to try many different models, covering many similarity assumptions, with many settings of complexity for each model, and choose the one according to some model selection criteria. In this paper, we take a nonparametric Bayesian approach to this problem (using the Dirichlet Process and the Indian Buffet Process as building blocks) such that the appropriate MTL model capturing the correct task relatedness structure and the model complexity for *that* model will be learned in a data-driven manner side-stepping the model selection issues.

### 2.1. The Dirichlet Process

The Dirichlet Process (DP) is a prior distribution over discrete distributions (Ferguson, 1973). Discreteness implies that if one draws samples from a distribution drawn from the DP, the samples will cluster: new samples take the same value as older samples with some positive probability. A DP is defined by two parameters: a concentration parameter $\alpha$ and a base measure $G_0$. The sampling process defining the DP draws the first sample from the base measure $G_0$. Each subsequent sample would take on a new value drawn from $G_0$ with a probability proportional to $\alpha$, or reuse a previously drawn value with probability proportional to the number of samples having that value. This property makes it suitable as a prior for effectively infinite mixture models, where the number of mixtures can grow as new samples are observed. Our mixture of factor analyzers based MTL model uses the DP to model the mixture components so we do not need to specify their number *a priori*.

### 2.2. The Indian Buffet Process

The Indian Buffet Process (IBP) (Griffiths & Ghahramani, 2006) and the closely related Beta Process (Thibaux & Jordan, 2007) define a distribution on a *collection* of sparse binary vectors of unbounded size (or, equivalently, on sparse binary matrices with one dimension fixed but the other being unbounded). Such sparse structures are commonly used in applications such as sparse factor analysis (Paisley & Carin, 2009) where we want to decompose a data matrix $X$ such that each observation $X_n \in \mathbb{R}^D$ is represented as a sparse combination of a set of $K \ll D$ basis vectors (or *factors*) but $K$ is not specified *a priori*. The generative story in the finite case is (assuming a linear Gaussian model generation):

$$\begin{aligned} X_n &\sim \mathcal{N}or(\Lambda b_n, \sigma_X^2 \mathbf{I}) \\ \Lambda_k &\sim \mathcal{N}or(0, \sigma^2 I) \\ b_{kn} &\sim \mathcal{B}er(\pi_k) \\ \pi_k &\sim \mathcal{B}et(\alpha/K, 1) \end{aligned}$$

In the above, $\Lambda$ is a matrix consisting of $K$ columns (the factors) and the factor combination is defined by the sparse binary vector $b_n$ of size $K$. For the more general case of factor analysis, factor combination weights are sparse real-valued vectors, so the model is of the form $X_n = \Lambda(s_n \odot b_n) + E$, where $s_n$ is a real-valued vector of the same size as $b_n$ (Paisley & Carin, 2009) and can be given a Gaussian prior, and $\odot$ is the elementwise product. Our mixture of factor analyzers based MTL model uses the IBP/Beta Process to model each factor analyzer so we do not need to specify the number of factors $K$ *a priori*.

## 3. Mixture of Factor Analyzers based Generative Model for MTL

Our proposed model assumes that the parameters (i.e., the weight vector) of each task are sampled from a mixture of factor analyzers (Ghahramani & Beal, 2000). Note that our model is defined over *latent* weight vectors whereas the standard mixture of factor analyzers is commonly defined to model *observed data*.

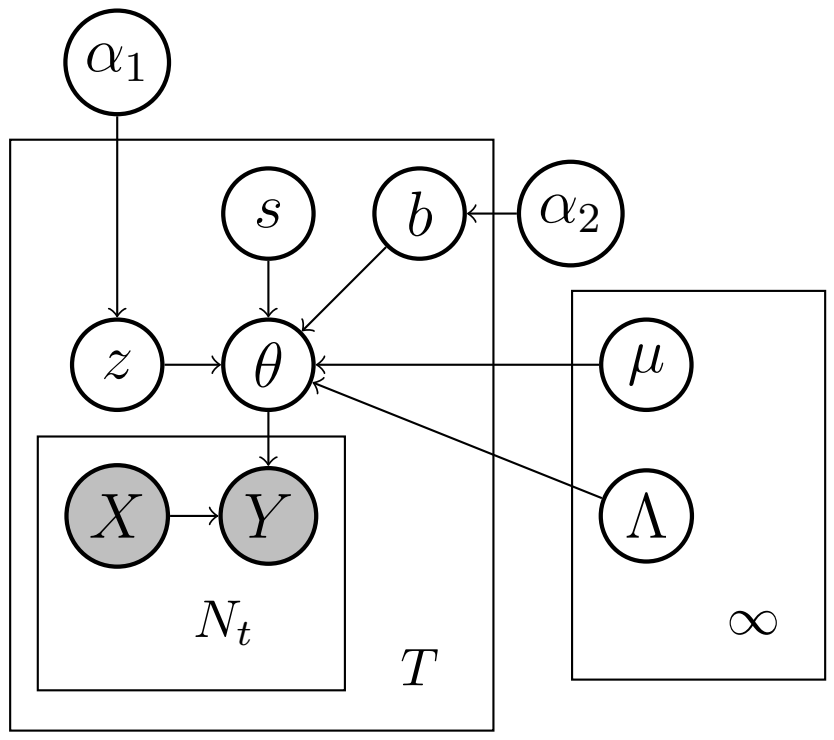

*Figure 1.* A graphical depiction of our model. The task parameters $\theta$ are sampled from a DP-IBP mixture and used to generate the $Y$ values.

We assume that we are learning $T$ related tasks, where each task is represented by a weight vector $\theta_t \in \mathbb{R}^D$ that is assumed to be sampled from a mixture of $F$ factor analyzers where each factor analyzer consists of $K \leq \min\{T, D\}$ factors (note: our model also allows each factor analyzer to have a different number of factors). Here $D$ denotes the number of features in the data. Each task is a set of $X$ and $Y$ values, and each $Y$ is assumed to be generated from the corresponding $X$ value and task weight vector. In our model, the weight vector $\theta_t$ for task $t$ is generated by first sampling a factor analyzer (defined by a mean task parameter $\mu_t \in \mathbb{R}^D$ and a factor loading matrix $\Lambda_t \in \mathbb{R}^{D\times K}$) using the DP, and then generating $\theta_t$ using *that* factor analyzer. In equations, this be written as $\theta_t = \mu_t + \Lambda_t f_t + \varepsilon_t$.

The weight vector $\theta_t$ is a *sparse* linear combination of $K$ *basis vectors* represented by the columns of $\Lambda_t$ (each column is a "basis task"). The combination weights are given by $f_t \in \mathbb{R}^K$ which we represent as $s_t \odot b_t$ where $s_t$ is a real valued vector and $b_t$ is a binary valued vector, both of size $K$. Our model uses a Beta-Bernoulli/IBP prior on $b_t$ to determine $K$, the number of factors in each factor analyzer. The $\{\mu_t, \Lambda_t\}$ pair for each task is drawn from a DP, also giving the tasks a clustering property, and there will be a finite number $F \leq T$ of *distinct* factor analyzers. Finally, $\varepsilon_t \sim \mathcal{N}or(0, \frac{1}{\sigma^2}\mathbf{I})$ represents task-specific noise.

Figure 1 shows a graphical depiction of our model and Figure 2 shows the generative story for the linear regression case . The DP base measure $G_0$ is a product of two Gaussian priors for $\mu_t, \Lambda_t$. In our nonparametric Bayesian model, $F$ and $K$ need not be known *a priori*; these are inferred from the data.

For classification, the only change is that the first line in the generative model becomes $Y_{t,i} \sim \mathcal{B}er(sig(\theta_t \cdot X_{t,i}))$, where $sig(x) = \frac{1}{1+\exp(-x)}$ is the logistic function and $\mathcal{B}er$ is the Bernoulli distribution.

$$
\begin{aligned}
Y_{t,i} &\sim \mathcal{N}or(\theta_t^T X_{t,i}, \mathbf{I}) \\
\theta_t &\sim \mathcal{N}or(\mu_t + \Lambda_t \cdot (s_t \odot b_t), \frac{1}{\sigma^2}\mathbf{I})) \\
\mu_t, \Lambda_t &\sim G \quad s_t \sim \mathcal{N}or(0, \mathbf{I}) \quad b_{kt} \sim \mathcal{B}er(\pi_k) \\
G &\sim \mathcal{DP}(\alpha_1, G_0) \quad \pi_k \sim \mathcal{B}et(\alpha_2/K, 1)
\end{aligned}
$$

*Figure 2.* The hierarchical model. The indicator variable $z$ of Fig 1 is implicit in the draw from the DP. The Beta-Bernoulli draw for $b_{kt}$ approximates the IBP for large $K$ (actual $K$ will be inferred from the data).

A number of existing multitask learning models arise as special cases of our model as it nicely interpolates between some different and useful scenarios, depending on the actual inferred values of $F$ and $K$, for a given multitask learning dataset:

- **Shared Gaussian Prior**($F$=1, $K$=0): (Chelba & Acero, 2006). This corresponds to a single factor analyzer modeling either a diagonal or full-rank Gaussian as the prior.
- **Cluster-based Assumption**($F > 1, K$=0): (Xue et al., 2007; Jacob & Bach, 2008). This corresponds to a mixture of identity-covariance or full-rank Gaussians as the prior.
- **Linear Subspace Assumption**($F$=1, $K < D$): (Zhang et al., 2006; Rai & Daumé III, 2010). This corresponds to a single factor analyzer with less than full rank. Note that this is also equivalent to the matrix $\Theta = \{\theta_1, \ldots, \theta_T\}$ being a rank-$K$ matrix (Argyriou et al., 2007).
- **Nonlinear Manifold Assumption**: A mixture of linear subspaces allows modeling a nonlinear subspace (Chen et al., 2010) and can capture the case when the weight vectors live on a nonlinear manifold (Ghosn & Bengio, 2003; Agarwal et al., 2010). Moreover, in our model, the manifold's intrinsic dimensionality can be different in different parts of the ambient space (since we do not restrict $K$ to be the same for each factor analyzer).

Our nonparametric Bayesian model can interpolate between these cases as appropriate for a given dataset, without changing the model structure or hyperparameters. From a non-probabilistic analogy, our model can be seen as doing dictionary learning/sparse coding (Aharon et al., 2010) over the *latent* weight vectors (albeit, using an *undercomplete* dictionary setting since we assume $K \leq \min\{T, D\}$). The model learns $M$ dictionaries of basis tasks (one dictionary per group/cluster of tasks, and $M$ inferred from the data) and tasks within each cluster are expressed as a sparse linear combination of elements from *that* dictionary. Our model can also be generalized further, e.g., by replacing the Gaussian prior on the low-dimensional latent task representations $s_t \in \mathbb{R}^K$ by a prior of the form $P(s_{t+1}|s_t)$, one can even relax the exchangeability assumption of tasks within each group, and have tasks that are evolving with time.

### 3.1. Variational inference

As this model is infinite and combinatorial in nature, exact inference is intractable and sampling-based inference may take too long to converge (Doshi-Velez et al., 2009; Blei & Jordan, 2006). Hence, we employ a variational mean-field algorithm to perform inference in this model. To do so, we lower-bound the marginal log-probability of $Y$ given $X$ using a fully factored approximating distribution $Q$ over the model parameters $\theta, \mu, \Lambda, z, b, s$:

$$\begin{aligned} \log P(Y|X) &= \log E_P[P(Y|X, \theta, \mu, \Lambda, z, b, s)] \\ &\geq E_Q[\log P(Y|X)] \\ &\quad - E_Q[\log Q(Y|X)]. \end{aligned}$$

To do so, we approximate the DP and the IBP with a tractable distribution $Q$. For the DP we use a finite stick-breaking distribution, based on the infinite stick-breaking representation of the DP (Blei & Jordan, 2006). In this representation, we introduce, for each $\theta_t$, a multinomial random variable $z_t$ that indexes the infinite set of possible mixture parameters $\mu$ and $\Lambda$. The $z_t$ vector is nonzero on its $i$-th component with probability $\phi_i \prod_{j<i}(1-\phi_j)$, where $\phi$ is an infinite set of independent $\mathcal{B}et(1, \alpha_1)$ random variables ($\mathcal{B}et$ is the Beta distribution). A finite approximation to the DP is obtained by setting a given $\phi_i$ to 1, which sets the probability of $z_j$ for $j > i$ necessarily to 0. While there is a similar stick-breaking construction to the IBP (Teh et al., 2007), it is not in the exponential family and requires complicated approximations, so we represent the IBP by its finite Beta-Bernoulli approximation (Doshi-Velez et al., 2009).

The distribution we are approximating then (for the linear regression case) is shown in Figure 3 (top). The stick-breaking distribution $SBP$ which is the prior for $z_t$ is such that $P(z_t = i) = \phi_i \prod_{j<i}(1-\phi_j)$.

In our variational distribution, we set the number of factor analyzers in the truncated stick-breaking representation to a hyperparameter $F$ and the number of factors in each such analyzer to a truncation level hyperparameter $K$. After inference, if the truncation levels are set high enough, most factor analyzers (and factors within each factor analyzer) will not be used, effectively approximating the property of the infinite model that only a small finite number of components is ever used to model a finite data set. It is worthwhile to note that while the solution found by the variational approximation is necessarily finite and with complexity bounded by the truncation parameters, it will still implicitly perform model selection. Therefore, more often than not, it will concentrate most of its posterior mass on models with less complexity than the truncation parameters suggest. Ishwaran & James (2001) present two theorems to help choose these truncation levels, as using smaller values of $F$ and $K$ (particularly $K$, as the update equations are quadratic in $K$) can lead to significant savings of computing time (in our experiments, we simply set these to $\min\{D, T\}$) which we found to be sufficient).

$$\begin{aligned} Y_{t,i} &\sim \mathcal{N}or(\theta_t^T X_{t,i}, \mathbf{I}). \\ \theta_t &\sim \mathcal{N}or(\mu_{z_t} + \Lambda_{z_t}(s_{t,z_t} \odot b_{t,z_t}), \frac{1}{\sigma^2}\mathbf{I}) \\ \mu_f &\sim \mathcal{N}or(0, \mathbf{I}), \quad \Lambda_{f,k} \sim \mathcal{N}or(0, \mathbf{I}) \\ s_{t,f} &\sim \mathcal{N}or(0, \mathbf{I}), \quad b_{t,f,k} \sim \mathcal{B}er(\beta_{f,k}) \\ z_t &\sim SBP(\phi), \quad \beta_{f,k} \sim \mathcal{B}et(\alpha_2/K, 1) \\ \phi_f &\sim \mathcal{B}et(1, \alpha_1) \end{aligned}$$

$$\begin{aligned} Q(\theta_t) &= \mathcal{N}or(\nu_{\theta_t}, \mathbf{I}) \\ Q(\mu_f) &= \mathcal{N}or(\nu_{\mu_f}, \mathbf{I}), \quad Q(\Lambda_f) = \mathcal{N}or(\nu_{\Lambda_f}, \mathbf{I}) \\ Q(s_{t,f}) &= \mathcal{N}or(\nu_{s_{t,f}}, \mathbf{I}), \quad Q(b) = \mathcal{B}er(\nu_b) \\ Q(z_t = i) &= \nu_{z_{t,i}}, \quad Q(\beta) = \mathcal{B}et(\rho_1, \rho_2) \\ Q(\phi) &= \mathcal{B}et(\gamma_1, \gamma_2) \end{aligned}$$

*Figure 3.* Top: the distribution being approximated. Bottom: Our approximating $Q$ distribution (note: $P(Y|\theta)$ is lower-bounded directly)

Our approximating $Q$ distribution is shown in Figure 3 (bottom). For the linear regression case, we treat $P(Y|\theta)$ by lower-bounding it directly, without introducing an approximating distribution for $Y$. In the case of logistic regression, we use the lower bound by (Jaakkola & Jordan, 1996) that allows us to integrate out the logistic function.

Apart from approximating the DP with the truncated stick-breaking prior, approximating the IBP with a set of symmetric, finite Beta distributed variables, and lower-bounding the logistic function with a quadratic, all the computations involved in deriving the variational lower bound are straightforward exponential-family computations. Note that for $Q$ we could use more general covariances instead of the identity matrices. In practice, we found that this did not improve classification performance, and it would imply on a significantly higher computational cost. Another less expensive option however would be to use the same hyperparameter for each feature, i.e., a spherical (instead of diagonal) covariance $\tau^2\mathbf{I}$ which would require optimizing w.r.t. a single hyperparameter $\tau$. The variational parameter updates are:

$$
\begin{aligned}
\gamma_{f,1} &= 1 + \sum_t \nu_{z_{t,f}} \\
\gamma_{f,2} &= \alpha_1 + \sum_t \sum_{j>f} \nu_{z_{t,j}} \\
\nu_{z_{t,f}} &\propto \exp\Big(\mathbf{\Psi}(\gamma_{f,1}) - \mathbf{\Psi}(\gamma_{f,1}+\gamma_{f,2}) \\
&\quad + \sum_{j<f} (\mathbf{\Psi}(\gamma_{j,2}) - \mathbf{\Psi}(\gamma_{j,1}+\gamma_{j,2})) \\
&\quad + E_Q[\log P(\theta_t | z_t = f)]\Big) \\
\rho_{f,k,1} &= \frac{\alpha_2}{K} + \sum_t \nu_{b_{t,f,k}}, \quad \rho_{f,k,2} = 1 + \sum_t (1 - \nu_{b_{t,f,k}}) \\
\nu_{b_{t,f,k}} &= sig\Big(\mathbf{\Psi}(\rho_{f,k,1}) - \mathbf{\Psi}(\rho_{f,k,2}) \\
&\quad + \sigma\nu_{z_{t,f}}\Big(\Big[\nu_{\theta_t} - \nu_{\mu_f} - (\nu_{s_{t,i}}+1)\nu_{\Lambda_{f,i}} \\
&\quad - \sum_{j\neq i} \nu_{s_{t,j}} \nu_{b_{t,f,j}} \nu_{\Lambda_{f,j}}\Big]^T \nu_{\Lambda_{f,i}} \nu_{s_{t,i}} \\
&\quad - \frac{D}{2}\nu^2_{s_{t,i}} - \frac{DF}{2}\Big)\Big) \\
\nu_{s_{t,i}} &= (1 + \sigma \nu_{z_{t,f}} \nu_{b_{t,f,i}} (D + ||\nu_{\Lambda_{f,i}}||^2))^{-1} \\
&\quad \nu_{z_{t,f}} \sigma \Big(\Big(\nu_{\theta_t} - \nu_{\mu_f} \\
&\quad - 0.5 \sum_{j\neq i} \nu_{s_{t,f,j}} \nu_{b_{t,f,j}} \nu_{\Lambda_{f,j}}\Big)^T \nu_{\Lambda_{f,i}} \nu_{b_{t,f,i}}\Big) \\
\nu_{\mu_f} &= \frac{\sum_t \nu_{z_{t,f}} \sigma(\nu_{\theta_t} - \nu_{\Lambda_f}(\nu_{s_{t,f}} \odot \nu_{b_{t,f}}))}{1 + \sigma \sum_t \nu_{z_{t,f}}} \\
\nu_{\Lambda_{f,i}} &= \Big(1 + \sigma \sum_t \nu_{z_{t,f}} \nu_{b_{t,f,i}} (1 + \nu^2_{s_{t,f,i}})\Big)^{-1} \\
&\quad \sigma \sum_t \nu_{z_{t,f}} \nu_{s_{t,f,i}} \nu_{b_{t,f,i}} \Big(\nu_{\theta_t} - \nu_{\mu_f} \\
&\quad - \frac{1}{2} \sum_{j\neq i} \nu_{s_{t,f,j}} \nu_{b_{t,f,j}} \nu_{\Lambda_{f,j}}\Big)
\end{aligned}
$$

In the above $\mathbf{\Psi}$ denotes the digamma function. While it is possible to update $\nu_{\theta_t}$ analytically, the update requires inverting a matrix, and in our experiment this matrix was often ill-conditioned, so we updated $\nu_{\theta_t}$ by optimizing the lower bound with the L-BFGS-B optimizer (Zhu et al., 1997). The optimizer is run until convergence at each iteration, warm-started with the previous value. We note that it could be replaced by any other optimizer, including gradient methods, with no changes in the above equations.

The complete derivations are provided in the the supplementary material.

For regression, the gradient of the lower bound with respect to $\nu_{\theta_t}$ is

$$
\begin{aligned}
\nabla L(\nu_{\theta_t}) &= \sigma \sum_f \nu_{z_{t,f}} \left(\nu_{\theta_t} - \nu_{\mu_f} - \nu_{\Lambda_f}(\nu_{s_{t,f}} \odot \nu_{b_{t,f}})\right) \\
&\quad + \sum_i^{N_t} \left(Y_{t,i} X_{t,i} - X_{t,i} X_{t,i}^T \nu_{\theta_t}\right).
\end{aligned}
$$

For classification the gradient is similar, the main difference being that there is an extra factor in the $X_{t,i}X_{t,i}^T\nu_{\theta_t}$ term involving the variational parameter for the lower bound of the logistic function.

We also optimize the lower bound w.r.t the precision parameter $\sigma$ to obtain an empirical Bayes estimate:

$$
\begin{aligned}
\frac{1}{\sigma} &= \sum_t \sum_f \nu_{z_{t,f}} \Bigg(\frac{||\nu_{\theta_t} - \nu_{\mu_f} - \nu_{\Lambda_f}(\nu_{s_{t,f}} \odot \nu_{b_{t,f}})||^2}{KDF} \\
&\quad + \frac{\sum_i \nu_{b_{t,f,i}} (\nu^2_{s_{t,f,i}} + ||\nu_{\Lambda_{f,i}}||^2)}{KF} + \frac{1}{K}\Bigg).
\end{aligned}
$$

The hyperparameters $\alpha_1$ and $\alpha_2$ are held fixed and can be optimized by cross-validation. We initialize the inference process with $\nu_{\theta_t}$ set to the maximum likelihood solution to each task's regression or classification problem. Then we alternate updating all other parameters to convergence and updating $\nu_{\theta_t}$ given the other parameters. The value of $\nu_{\theta_t}$, and hence the regression or classification accuracy, usually stabilizes after the first couple of iterations, and the only changes observed are further improvements to the lower bound. This matches behavior observed in Ando & Zhang (2005). All our experiments were run on three iterations.

## 4. Experiments

We present results on both synthetic and real-world datasets, and on linear regression and classification settings. As a sanity check to show that our model can learn the underlying latent task structures correctly, we generated a synthetic data consisting of 5 clusters of tasks. Each cluster consists of 10 binary classification tasks, having 100 examples each. We used a 50/50 split for train/test data. Each task is represented by a weight vector of length $D = 20$. Figure 4 (left) shows the true correlation structure of the tasks and Figure 4 (right) shows the recovered structure by our model: it correctly infers the correct number (5) of clusters. Our model resulted in a classification accuracy of 83.2%, whereas independently learned tasks resulted in an accuracy of 79.2%.

Our next set of experiments compare our model with a number of baseline methods on several synthetic and real-world multitask regression and multitask classification problems. Our baselines include:

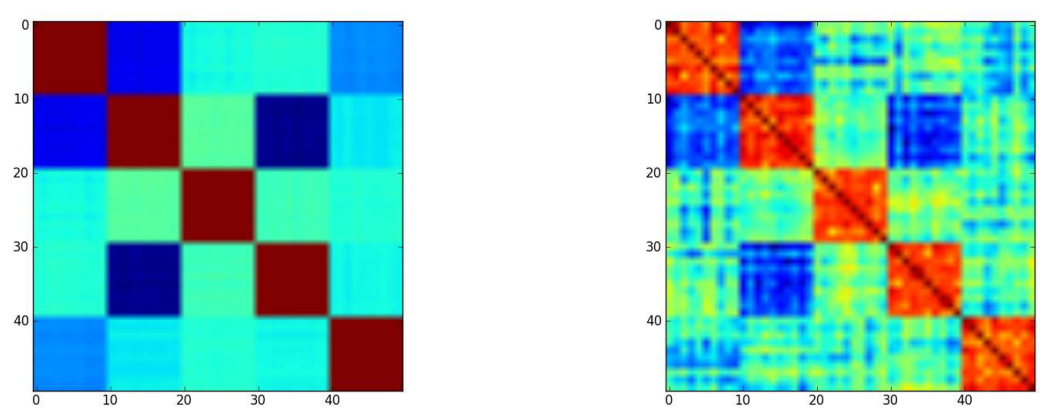

*Figure 4.* Left: Plot of the correlation matrix of the ground-truth weight vectors of the 50 tasks. Right: Inferred correlation matrix

- Independently learned tasks - **STL**: assumes the tasks are independent (no information sharing).
- Multitask Feature Learning - **MTFL**: assumes the tasks share a common set of features (Argyriou et al., 2007).
- Shared Gaussian prior over the weight vectors - **PRIOR** (Chelba & Acero, 2006): assumes the tasks are drawn from a shared Gaussian prior with a unknown but fixed mean and covariance.
- Single shared subspace - **RANK** (Zhang et al., 2006; Rai & Daumé III, 2010): assumes the tasks live close to a linear subspace (also equivalent to the matrix of the weight vector being low-rank).
- DP mixture model based task clustering - **DP-MTL** (Xue et al., 2007): assumes the weight vectors are generated from a mixture model, each component being a full-rank Gaussian.
- **L**earning with **W**hom to **S**hare - **LWS** (Kang et al., 2011). It is an integer programming based method that learn the task grouping structure (with pre-specified number of groups) and encourages the tasks within each group to share features.

Of these baselines, MTFL and LWS were used for regression problems only since the publicly available implementations are for regression. In the experiments, we would refer to our model as **MFA-MTL** (**M**ixture of **F**actor **A**nalyzers for **M**ulti**T**ask **L**earning). In all our experiments, we set the hyperparameters $\alpha_1 = 1$ and $\alpha_2 = 5$, as these values performed reasonably in preliminary experiments. The truncation level for the DP can be chosen to be equal to the number of tasks $T$, and for the IBP, to be the minimum of $T$ and the number of features $D$ in the data. This is often more than necessary and in most of our experiments, much smaller truncation levels were found to be sufficient.

For our multitask regression experiments, we compared MFA-MTL with STL, MTFL, and LWS (we skip the other baselines as they performed comparably or worse than MTFL/LWS). For this experiment, we used three datasets - one synthetic dataset used in (Kang et al., 2011), and two real-world datasets used commonly in the multitask learning literature: (1) **School**: This dataset consists of the examination scores of 15362 students from 139 schools in London. Each school is a task so there are a total of 139 tasks for this dataset. (2) **Computer**: This dataset consists of a survey of 190 students about the chances of purchasing 20 different personal computers. There are a total of 190 tasks, 20 examples per task, and 13 features per example. For the synthetic data, we followed the similar procedure for train/test split as used by (Kang et al., 2011). For School and Computer datasets, we split the data equally into training and test set and further only used 20% of the training data (training set deliberately kept small as is often the case with multitask learning problems in practice). The average mean squared errors (i.e., across tasks) in predicting the responses by each method are shown in Table 1. As shown in Table 1, MFA-MTL outperforms the other baselines on all the datasets. Moreover, for the synthetic data, we found that it also inferred the number of task groups (3) correctly (the LWS baseline needs this number to be specified - we ran it with the ground truth). On the school and computer datasets, MFA-MTL outperforms STL and LWS and does slightly better than MTFL. For LWS on these two datasets, we report the best results as obtained by varying the number of groups from 1 to 20.

| | **Synthetic** | **School** | **Computer** |
|---|---|---|---|
| STL | 1.35 | 468.7 | 153.3 |
| MTFL | 0.36 | 376.1 | 30.4 |
| LWS | 0.37 | 430.9 | 30.2 |
| MFA-MTL | **0.18** | **374.5** | **29.8** |

*Table 1.* Mean squared error (MSE) of various methods on multitask regression problems

| | **Landmine** | **20ng** |
|---|---|---|
| STL | 52.9% | 69.3% |
| PRIOR | 52.9% | 75.8% |
| RANK | 53.8% | 75.8% |
| DP-MTL | 53.8% | 75.7% |
| MFA-MTL | **62.4%** | **76.9%** |

*Table 2.* Multitask classification accuracies of various methods on the **Landmine** and **20ng** datasets

We next experiment with the classification setting. For this, we chose two datasets: (1) **Landmine**: The landmine detection dataset is a subset of the dataset used in the symmetric multitask learning experiment by (Xue et al., 2007). It contains 19 classification tasks and the tasks are known to be clustered for this data. (2) **20ng:** We did the standard training/test split of 20 Newsgroups for multitask learning, following Raina et al. (2006), and used a 50/50 split for the landmine data. The classification accuracies reported by our model and the various baselines on landmine and 20 Newsgroups datasets are shown in Table 2. As shown in Table 2, our method outperforms the various baselines. We note that 3 of them (PRIOR, RANK, and DP-MTL), which are methods proposed in prior work, are special cases of our model (as discussed in Section 3). In particular, RANK performs worse than our method, potentially because all weight vectors share the same subspace which may not be desirable if not all the tasks are related with each other. DP-MTL performs worse than our method, potentially because it fits a *full-rank* Gaussian for each mixture component and is especially prone to overfit if the number of tasks is smaller than the number of features.

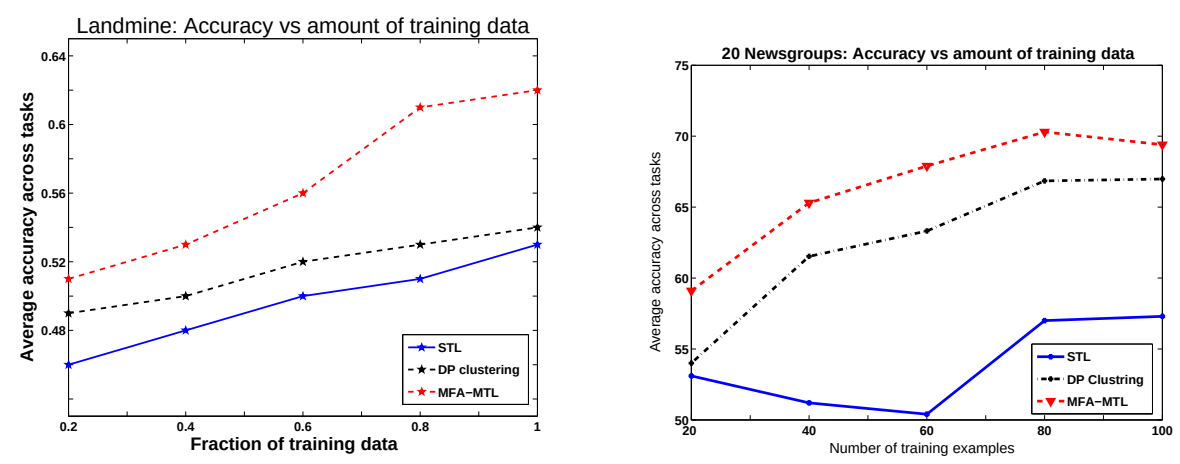

*Figure 5.* Average accuracies w.r.t. varying amount of training data (left: landmine data, right: 20ng data).

Finally, we investigated the behavior of different algorithms in the small training data regimes. For this, we varied the amount of training examples per task (for landmine data, we varied the fraction from 20% to 100%; for 20 Newsgroup, we varied the number of examples from 20 to 100). Results are shown in Figure 5. To uncrowd the figure, we compare only with STL and DP-MTL (the best performing baseline). In the small data regimes, our algorithm performs better as compared to both STL and DP-MTL. Another important aspect of an MTL algorithm is its asymptotic behavior in the limit of large training data per task. For this experiment, we compared MFA-MTL with STL on the school multitask regression dataset by providing each algorithm the complete training data. MFA-MTL resulted in an MSE of 261.4 as compared to STL which gave an MSE of 271.1. Therefore our algorithm tends to do comparably (in fact, marginally better) to independently learned tasks even when the amount of training data per task is sufficiently large.

## 5. Related Work

Apart from the prior work on multitask learning discussed in Section 1, our model is based on a somewhat similar motivation as the model proposed in (Argyriou et al., 2008). Their model assumes that tasks can be partitioned into groups and tasks within each group share a kernel. Their assumption is an extension of the earlier work on Multitask Feature Learning (Argyriou et al., 2007) (one of the baselines we used in our experiments) that assumes all tasks share the common kernel. In (Kumar & Daumé III, 2012), the authors assume that there is *single* set of task *basis vectors* (i.e., a task dictionary) and each task is a *sparse* combination of these basis vectors. In their model, the number of basis vectors shared between two tasks (i.e., their "overlap") can be seen as the pairwise task similarity. In Kang et al. (2011), the authors proposed a model based on the assumption that the tasks exist in groups and the tasks within each group share features, which is again similar in spirit to our work (this model was one of our baselines in the experiments). In contrast, the generative model we presented in this paper offers a number of advantages over these models such as the ability to deal with missing data in a principled manner, doing automatic model complexity control in a nonparametric Bayesian setting, and being flexible enough to subsume these and many other notions as task relatedness used in multitask learning.

Among other related work, Canini et al. (2010) propose hierarchical Dirichlet process models as good models for human categorical learning. The idea is that one can model transfer learning by assuming that people unsupervisedly learn subgroups of known classes and use these groups to refine the knowledge of new classes by sharing subgroups via a hierarchical Dirichlet process. Our model can be seen as a discriminative analog of their generative model, where aspects of the task parameter—instead of the distribution of the test examples—are shared among similar tasks and the sharing structure is discovered automatically.

## 6. Future Work and Discussion

We proposed and evaluated a nonparametric Bayesian multitask learning model that usefully interpolates between many different previously proposed models for estimating task parameters of multiple related learning problems, such as a shared Gaussian prior (Chelba & Acero, 2006), a clustering structure (Xue et al., 2007), reduced dimensionality (Argyriou et al., 2007; Zhang et al., 2006), manifold structure (Ghosn & Bengio, 2003; Agarwal et al., 2010), etc. We presented a variational mean-field algorithm for this model that exhibits competitive results on a set of synthetic as well as real-world multitask learning datasets. The proposed model, by using the flexibility afforded by nonparametric Bayesian techniques, requires only minimal assumptions to be applied to any given multitask learning problem. A possible future work is studying a hierarchical Dirichlet process variant of this model where different tasks are allowed to share exactly the *same* $\theta$ parameters, which might be beneficial in cases where training data is especially sparse or the tasks are more strongly clustered.

## Acknowledgments

The authors would like to thank Emma Tosch for helpful comments in the preparation of this manuscript. Piyush Rai was support by DARPA CSSG Grant P-1076-113840. Alexandre Passos was supported in part by the CIIR, in part by IARPA via DoI/NBC contract #D11PC20152, in part by UPenn NSF medium IIS-080384, in part by DARPA) Machine Reading Program under AFRL prime contract no. FA8750-09-C-0181. The U.S. Government is authorized to reproduce and distribute reprint for Governmental purposes notwithstanding any copyright annotation thereon. Any opinions, findings, and conclusion or recommendations expressed in this material are the authors' and do not necessarily reflect those of the sponsors.